\documentclass{article}
\usepackage{ijcai25}

\usepackage{times}
\usepackage{soul}
\usepackage{url}
\usepackage[hidelinks]{hyperref}
\usepackage[utf8]{inputenc}
\usepackage[small]{caption}
\usepackage{graphicx}
\usepackage{amsmath}
\usepackage{amsthm}
\usepackage{booktabs}
\usepackage{algorithm}
\usepackage{algorithmic}
\usepackage[switch]{lineno}

\usepackage{algorithm}
\usepackage{algorithmic}
\usepackage{subcaption}
\usepackage{booktabs}
\usepackage{array}
\usepackage{amssymb} 
\usepackage{amsthm} % 用于 Lemma 和 Proof 环境
\usepackage{amsmath}
\usepackage{enumitem}
\usepackage{autobreak}
\usepackage{breqn}
\usepackage{arydshln} % 引入虚线表格包
\usepackage{booktabs} % 引入高质量表格线

\title{Graph Random Walk with Feature-Label Space Alignment: A Multi-Label Feature Selection Method}

\author{
    Wanfu Gao\textsuperscript{1,2} \and 
    Jun Gao\textsuperscript{1,2}\thanks{Corresponding author} \and 
    Qingqi Han\textsuperscript{1,2} \and 
    Hanlin Pan\textsuperscript{1,2} \And 
    Kunpeng Liu\textsuperscript{3} 
     \\
    \affiliations{
        \textsuperscript{1}College of Computer Science and Technology, Jilin University, China\\
        \textsuperscript{2}Key Laboratory of Symbolic Computation and Knowledge Engineering of Ministry of Education, Jilin
University, China\\
        \textsuperscript{3}Department of Computer Science, Portland State University, Portland, OR 97201 USA
    }
    \emails
    gaowf@jlu.edu.cn, gaocheng23@mails.jlu.edu.cn, hanqq22@mails.jlu.edu.cn, panhl23@mails.jlu.edu.cn, kunpeng@pdx.edu
}

\begin{document}

\maketitle

\begin{abstract}
The rapid growth in feature dimension may introduce implicit associations between features and labels in multi-label datasets, making the relationships between features and labels increasingly complex. Moreover, existing methods often adopt low-dimensional linear decomposition to explore the associations between features and labels. However, linear decomposition struggles to capture complex nonlinear associations and may lead to misalignment between the feature space and the label space. To address these two critical challenges, we propose innovative solutions. First, we design a random walk graph that integrates feature-feature, label-label, and feature-label relationships to accurately capture nonlinear and implicit indirect associations, while optimizing the latent representations of associations between features and labels after low-rank decomposition. Second, we align the variable spaces by leveraging low-dimensional representation coefficients, while preserving the manifold structure between the original high-dimensional multi-label data and the low-dimensional representation space. Extensive experiments and ablation studies conducted on seven benchmark datasets and three representative datasets using various evaluation metrics demonstrate the superiority of the proposed method\footnote{Code: https://github.com/Heilong623/-GRW-}. 
\end{abstract}

\begin{figure}[t!]
    \centering
    \includegraphics[width=0.33\textwidth]{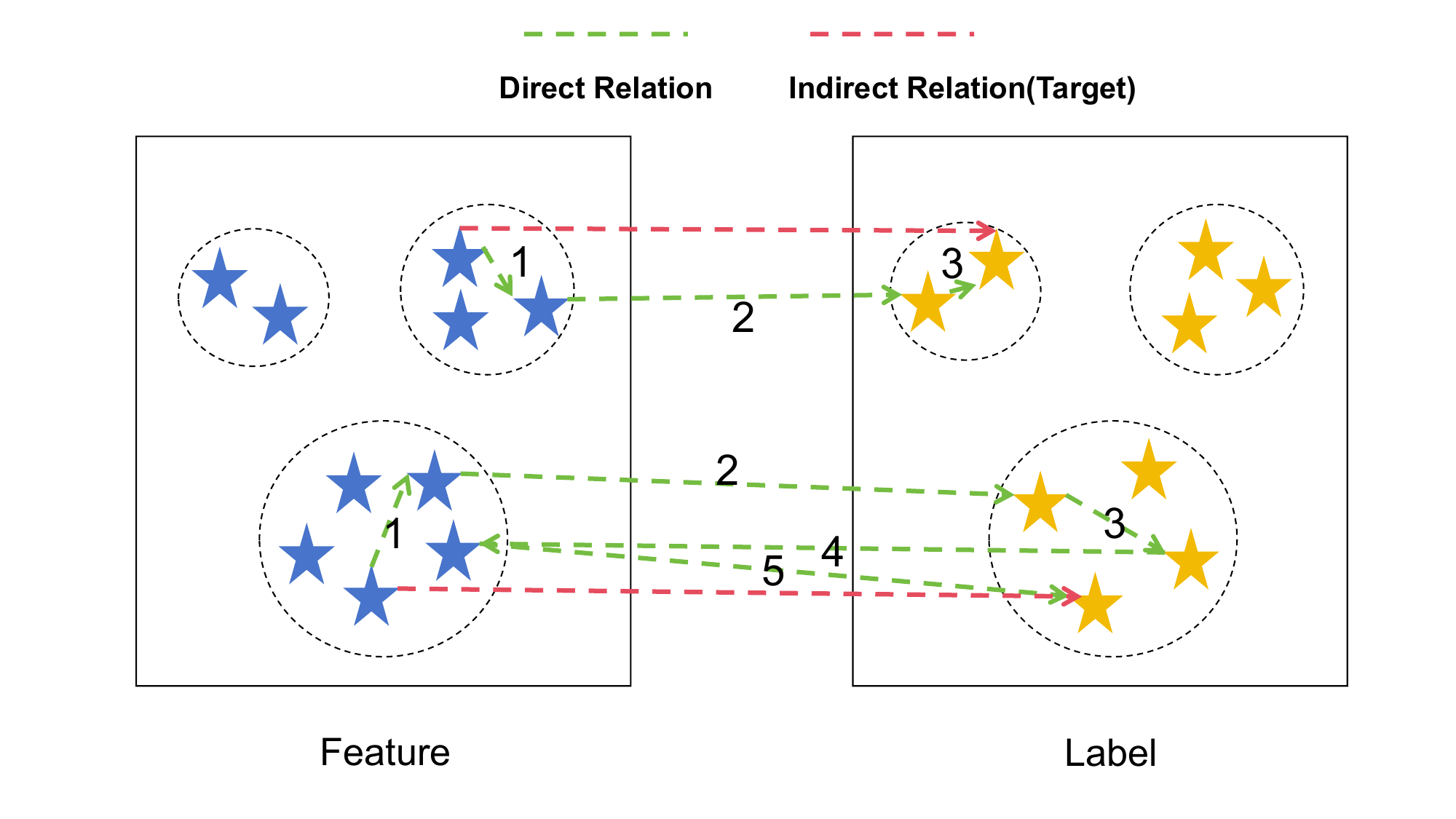}
    \caption{The diagram illustrates the direct and indirect associations between feature subsets and label subsets. For example, the underlying indirect relationships between features and labels can be captured through the direct relationships of 1, 2 and 3.}
    \label{fig:dictionary_learning}
\end{figure}
\section{Introduction}

In recent years, with the exponential growth of feature dimension, feature selection in high-dimensional datasets has become a critical task. By eliminating irrelevant and redundant features while retaining important and relevant ones, feature selection effectively reduces computational time and storage costs \cite{li2017feature}. As a result, feature selection methods have been widely applied across various domains,  including text mining \cite{jin2023feature}, image annotation \cite{kong2012multi}, music classification \cite{silla2008feature}, microarray data analysis \cite{bommert2022benchmark}, and biomarker discovery in proteomics \cite{tang2021metafs}. Generally, feature selection methods can be categorized into three main types: filter-based methods, wrapper-based methods, and embedded methods \cite{yao2017lle}.

Filter-based methods operate independently of the learning method \cite{zhang2021conditional}. In contrast, wrapper-based methods \cite{al2022wrapper} iteratively evaluate and select subsets of features to identify the optimal feature subset. Embedded methods, on the other hand, integrate the feature selection process directly into the learning method \cite{yamada2020feature}. Among embedded methods, many methods leverage sparse regularization to impose constraints that enhance the structural information of the data, thereby capturing the underlying structure more accurately and improving the generalization ability of the method. However, a key challenge for feature selection methods in multi-label learning problems is computing the relevance between features and labels.

The goal of multi-label learning is to assign multiple correct labels to a given instance \cite{zhu2018multi}, which poses new challenges compared to traditional single-label learning. Some studies address multi-label problems by transforming them into multiple single-label problems \cite{lee2015mutual,lee2017scls}, but these methods often overlook the correlations between labels. Other methods handle multi-label problems by creating label sets \cite{wang2021active}, grouping related labels together and converting the multi-label problem into a multi-class problem. However, these methods struggle to accurately capture the complex high-order correlations between features and labels in multi-label datasets.

In the real-world scenarios, a label can be associated with multiple other labels, making it inadequate to consider only pairwise label correlations \cite{zhang2021feature,lim2020mfc}. Consequently, the complex and deep relationships between labels have become a key focus of research \cite{yu2021multilabel,li2024multi}. Traditional multi-label feature selection methods primarily focus on the internal correlations within the feature set and the label set. Although some methods attempt to consider the associations between features and labels, this consideration remains insufficient \cite{wang2020icmsc}. For instance, linear decomposition methods such as Non-negative Matrix Factorization (NMF) may lead to the loss of nonlinear relationships within the data. This issue becomes more pronounced in real-world scenarios with complex dependencies. In the constructed composite diagram shown in Figure 1, indirect relationships between features and labels are established through the propagation of relevant nodes. Our method is specifically designed to focus on capturing these indirect relationships. To more comprehensively capture the nonlinear and indirect relationships between subsets of features and subsets of labels, we propose a novel multi-label feature selection method. The main contributions of this work are as follows:

\begin{itemize}
    \item We propose a random walk strategy on a feature-label composite graph constructed using mutual information, which captures both direct and implicit indirect associations between features and labels, thereby facilitating a comprehensive representation of the high-dimensional structures within the data.
    \item By aligning the low-rank representations of similar samples in the feature space and label space, we ensure that the shared latent space after low-rank decomposition effectively captures the structural consistency between the two spaces.
    \item Extensive experiments conducted on seven datasets validate the effectiveness of the proposed method.
\end{itemize}

\section{Related Work}

In the field of multi-label feature selection, numerous methods have been proposed to capture the correlations between features and labels \cite{sun2021feature}. With the increasing dimensionality of data, the diversity of labels also rises. Based on the depth of consideration given to label correlations, these methods are typically categorized into three types \cite{liu2023multi}, first-order strategies, second-order strategies, and high-order strategies.

First-order strategies address multi-label problems using single-label methods but ignore the correlations between labels \cite{lee2015mutual,lin2015multi}. Although this method simplifies the modeling process, it fails to capture label dependencies, making it challenging to identify key features within label relationships. Second-order strategies address this limitation to some extent by modeling pairwise relationships between labels \cite{xiong2021feature,qian2022label}. However, since label relationships often extend beyond simple pairwise connections, these methods only capture the associations between features and a subset of labels, falling short of comprehensively reflecting the complex dependencies between features and all labels.

In contrast, high-order strategies \cite{wu2020multi,huang2016learning} go further by considering the correlations between features and all possible label subsets, thus providing a more comprehensive modeling of the complex dependencies among labels. However, these methods focus on uncovering the direct dependencies between variables and fail to effectively capture implicit indirect correlations, which limits their ability to characterize deep, latent relationships.

Manifold learning \cite{fan2021manifold,cohen2023few} has been employed to ensure the consistency between feature correlation and label correlation. This method assumes that samples with similar features should have similar labels, thereby achieving alignment between feature and label similarity. The core idea is to uncover the intrinsic structure of high-dimensional data through its low-dimensional latent representations. \cite{tang2019unsupervised} proposed a graph-based manifold regularization method to capture the manifold structure of the data, applying the resulting latent representations to unsupervised feature selection. However, these methods fail to adequately address the alignment problem between the low-dimensional representations of the feature space and the label space.

In recent years, random walk techniques have seen increasing exploration and application in the field of feature selection. The Random Walk Feature Selection (RWFS) method \cite{feng2017novel} is based on the Optimal Feature Selection (OPFS) technique, which identifies feature subsets and integrates random walk methods with predefined thresholds to filter out redundant features. The Pattern-Based Local Random Walk (PBLRW) method \cite{song2019local} constructs local random walk models through feature combinations and assigns transition probabilities based on the strength of these combinations. However, the random walk operations in these methods remain limited to the feature space, neglecting label dependencies, and fail to fully leverage their capacity for capturing higher-order relationships.

\section{The Proposed Method}
We propose a multi-label feature selection method based on Graph Random Walk and Structured Correlation Matrix Factorization (GRW-SCMF). By incorporating graph random walk, the method captures both the direct correlations and implicit indirect correlations between features and labels. Furthermore, in the shared low-dimensional representation, it aligns similar samples in the feature space and label space to ensure consistency. The framework is shown in Figure 2.
\subsection*{Random Walk Graph}
In this paper, we construct a composite graph \( G = (V, E) \), where the vertex set \( V \) consists of two types of vertices: feature vertices \( V_f \) and label vertices \( V_l \). The edge set \( E \) includes the following three types of edges:
\begin{enumerate}
    \item Feature-feature edges \( E_{ff} \),
    \item Label-label edges \( E_{ll} \),
    \item Feature-label edges \( E_{fl} \).
\end{enumerate}

The adjacency matrices of the feature graph \( A_{\text{features}} \) and the label graph \( A_{\text{labels}} \) are calculated using the Gaussian kernel as follows:  

\begin{equation}
\small
A_{\text{features}}(i,j) = \exp\!\left(-\frac{\bigl\lVert X_{:,i} - X_{:,j} \bigr\rVert^{2}}{2\sigma^{2}}\right),
\end{equation}

\begin{equation}
\small
A_{\text{labels}}(i,j) = \exp\!\left(-\frac{\bigl\lVert Y_{:,i} - Y_{:,j} \bigr\rVert^{2}}{2\sigma^{2}}\right).
\end{equation}

Here, \( \|X_{:,i} - X_{:,j}\|^2 \) and \( \|Y_{:,i} - Y_{:,j}\|^2 \) represent the Euclidean distances between features and labels, respectively, and \( \sigma \) is the scale parameter of the Gaussian kernel.

The association between features and labels is represented by the mutual information matrix \( MI \), which is defined as follows:  

\begin{equation}
\small
MI(i,j) = \sum_{x,y} p(x,y) \log\left(\frac{p(x,y)}{p(x)p(y)}\right),
\end{equation}
where \( p(x,y) \) denotes the joint probability distribution of feature \( X \) and label \( Y \), and \( p(x) \) and \( p(y) \) represent the marginal probability distributions of \( X \) and \( Y \), respectively.  

To construct the transition probability matrices \( P_{\text{features}} \), \( P_{\text{labels}} \), and \( P_{lf} \) (the transpose of which is \( P_{fl} \)), we normalize the adjacency matrices of the feature graph \( A_{\text{features}} \) and label graph \( A_{\text{labels}} \), as well as the connection matrix \( MI \) between features and labels.

\subsection*{Random Walk Method}  
The random walk starts from a randomly selected feature vertex \( v \in V_f \) and updates its state according to the following rules:  

1. \textbf{When the current node is a feature node (\( v \in V_f \)):}  
   \begin{itemize}  
      \item With a probability of \( p_{\text{jump}} \), it jumps to a label node \( u \in V_l \). The target node is determined based on the transition probability \( P_{fl}(v,u) \), which is calculated as follows:  
      \begin{equation}
       P(v \to u) = p_{\text{jump}} \cdot P_{fl}(v,u), \quad u \in V_l.
      \end{equation}
      \item Otherwise, with a probability of \( 1 - p_{\text{jump}} \), it jumps to another feature node \( u \in V_f \). The target node is determined based on the transition probability \( P_{\text{features}}(v,u) \), which is calculated as follows:  
      \begin{equation}
      P(v \to u) = (1 - p_{\text{jump}}) \cdot P_{\text{features}}(v,u), \quad u \in V_f.
      \end{equation}
   \end{itemize}

2. \textbf{When the current node is a label node (\( v \in V_l \)):}  
   \begin{itemize}  
      \item With a probability of \( p_{\text{jump}} \), it jumps to a feature node \( u \in V_f \). The target node is determined based on the transition probability \( P_{lf}(v,u) \), which is calculated as follows:  
      \begin{equation}
      P(v \to u) = p_{\text{jump}} \cdot P_{lf}(v,u), \quad u \in V_f.
      \end{equation}  
      \item Otherwise, with a probability of \( 1 - p_{\text{jump}} \), it jumps to another label node \( u \in V_l \). The target node is determined based on the transition probability \( P_{\text{labels}}(v,u) \), which is calculated as follows:  
      \begin{equation}
      P(v \to u) = (1 - p_{\text{jump}}) \cdot P_{\text{labels}}(v,u), \quad u \in V_l.
      \end{equation}  
   \end{itemize}

\subsection*{Feature-Label Relationship Update Rule}  
Implicit indirect relationships refer to the potential high-order associations between feature nodes and label nodes that are transmitted through intermediate nodes (other features and labels). These relationships are difficult to capture using traditional linear decomposition methods. To address this issue, RWMI leverages random walks to dynamically generate interaction sequences between feature and label nodes, effectively capturing such indirect relationships.  

Specifically, during each random walk, we record each pair of feature nodes \( f \in V_f \) and label nodes \( l \in V_l \), and calculate their distance \( d(f, l) \) in the walk sequence, which represents the number of steps separating them. The association weight \( RW(f, l) \) is updated according to the following formula:  

\begin{equation}
RW(f, l) += \text{decay\_factor}^{d(f, l)} \cdot MI(f, l),
\end{equation}
where \( \text{decay\_factor} \) is a distance decay factor that controls the influence of the number of steps on the weight, and \( MI(f, l) \) is the mutual information between feature \( f \) and label \( l \), which mitigates uncertainties caused by randomness, ensuring that the results of the \( RW \) matrix are more accurate and robust.  

Through multiple random walks, RW not only captures the direct associations between features and labels but also computes implicit indirect relationships via the intermediate nodes in the walk sequences. This process constructs a feature-label association matrix that represents high-order associations, which is further scaled to the \([0,1]\) range via Min–Max normalization.

\begin{figure}[t!]
    \centering
    \includegraphics[width=0.48\textwidth]{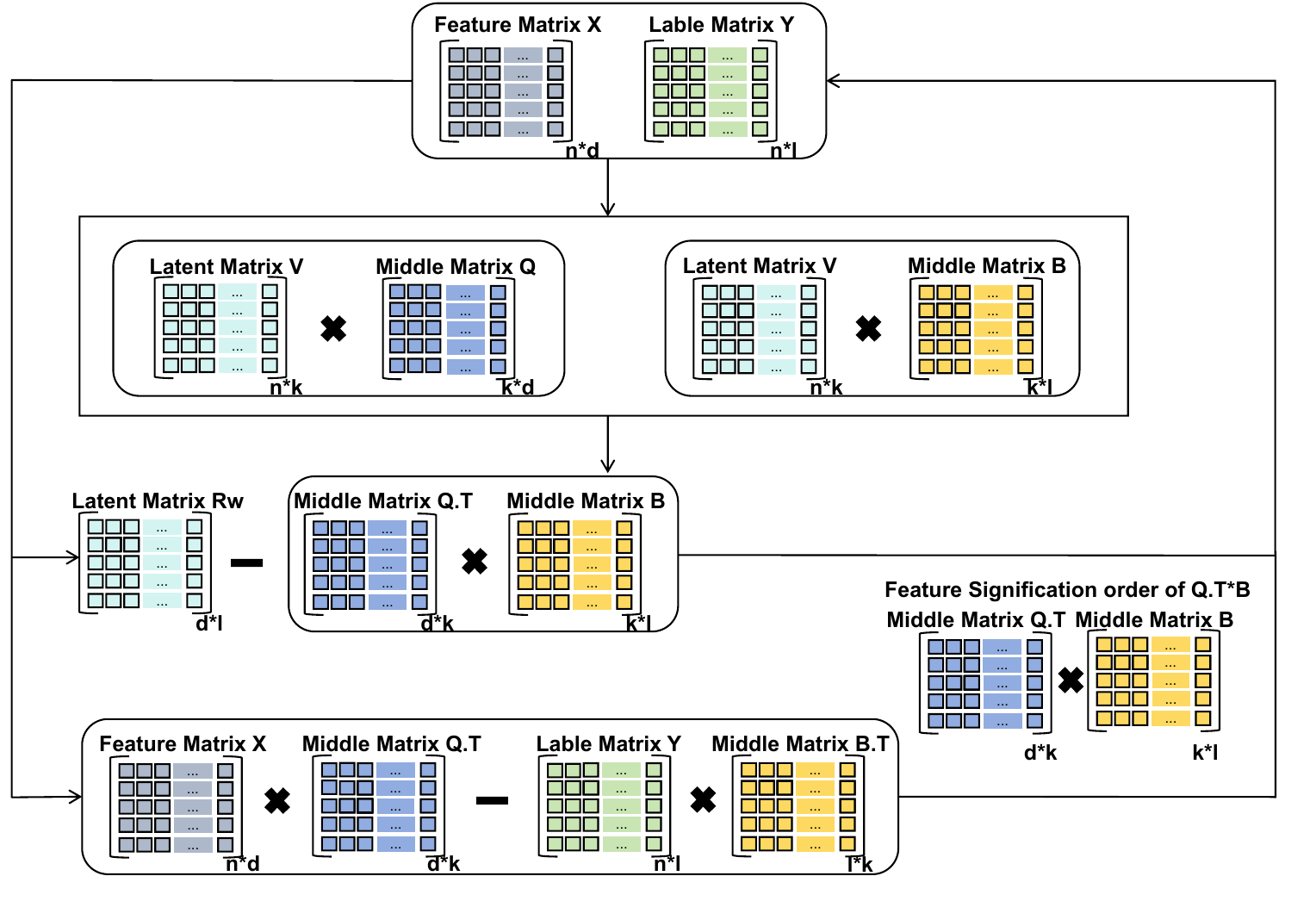}
    \caption{Algorithm framework. First, $X$ and $Y$ are decomposed using low-rank matrix factorization. Next, the feature-label association matrix is constrained by the random walk matrix $R_w$. Then, $X$ and $Y$ are aligned in a shared space. Finally, feature importance is computed via $Q^\top B$.}
    \label{fig:dictionary_learning}
\end{figure}

The procedure is detailed in Algorithm 1.

\subsection*{Objective Function}  
In multi-label data, the diversity of labels may introduce inaccurate or noisy label information, making it prone to incorporating irrelevant and redundant information when directly utilizing the original feature and label matrices for feature selection. To address this issue, we learned a shared low-dimensional latent semantic space for both the feature and label spaces. This low-dimensional space matrix condenses the critical information from features and labels while effectively reducing the influence of redundancy and noise. The objective function can be written as:

\begin{equation}
\small
\begin{aligned}
\min_{V, Q, B} &\ \|X - VQ\|_F^2 + \|Y - VB\|_F^2 \\
\end{aligned}
\end{equation}
\normalsize

The feature matrix \( X \in \mathbb{R}^{n \times d} \) represents the high-dimensional feature space of the samples, while the multi-label matrix \( Y \in \mathbb{R}^{n \times c} \) describes the label distribution of the samples. To uncover the shared distribution structure between the feature space and the label space, we introduce a low-dimensional latent semantic matrix \( V \in \mathbb{R}^{n \times k} \), which captures the low-dimensional shared representations of both spaces. The matrices \( Q \in \mathbb{R}^{k \times d} \) and \( B \in \mathbb{R}^{k \times c} \) are the low-dimensional representation coefficient matrices for features and labels, respectively, mapping the original data into the shared representation \( V \). Specifically, matrix \( V \) encodes the embedding relationships of features and labels in \( k \) latent semantic clusters. By applying non-negative matrix factorization (NMF) on \( Y \), the \( c \) labels can be effectively clustered, revealing the complex internal distribution structure of multi-label data.

\begin{algorithm}[t]
\small
\caption{Simplified Random Walk Mutual Information (RWMI)}
\begin{algorithmic}[1]
\REQUIRE $X$ (features), $Y$ (labels), $n\_walks$, $walk\_length$, $jump\_prob$, $decay\_factor$
\ENSURE Normalized RWMI matrix
\STATE Construct adjacency matrices $A_{features}$, $A_{labels}$ from $X$, $Y$.
\STATE Calculate and normalize mutual information matrix $MI$ between $X$ and $Y$.
\STATE Compute transition probability matrices: $P_{features}$, $P_{labels}$, $P_{feature\_to\_label}$, $P_{label\_to\_feature}$.
\STATE Initialize $RWMI$ matrix.
\FOR{$i = 1$ to $n\_walks$}
    \STATE Start random walk from a node.
    \FOR{$j = 1$ to $walk\_length$}
        \STATE Move between feature and label nodes based on probabilities and $jump\_prob$.
    \ENDFOR
    \STATE Update $RWMI$ matrix using decay factor and steps between feature-label pairs, and normalize it afterward.
\ENDFOR
\end{algorithmic}
\end{algorithm}

\begin{equation}
\small
\begin{aligned}
\min_{V, Q, B} \ &  \|X - VQ\|_F^2 +  \|Y - VB\|_F^2 +  \|R_w - Q^T B\|_F^2
\end{aligned}
\end{equation}
\normalsize

The term \( \|R_w - Q^T B\|^2_F \) introduced in the objective function (10) enhances the model's ability to capture the complex relationships between features and labels. Here, the matrix \( R_w \) is the feature-label implicit association matrix computed using the Random Walk Mutual Information (RWMI) method, which effectively captures indirect relations and latent dependencies in high-dimensional data. In contrast, \( Q^T B \) is the feature-label association matrix learned based on the low-dimensional shared semantic space. As it is derived from low-dimensional linear decomposition, it primarily captures linear relationships and direct associations. By combining the explicit modeling capability of linear decomposition with the implicit relationship mining power of random walks, this term significantly improves the model's ability to represent high-order nonlinear relationships between features and labels. Furthermore, this term ensures that the low-dimensional representation space preserves the manifold structure of the original high-dimensional data, aligning with the latest concepts in manifold learning, which posit that high-dimensional data often lie on a low-dimensional manifold.

\begin{equation}
\small
{
\begin{aligned}
\min_{V, Q, B} \ &  \|X - VQ\|_F^2 +  \|Y - VB\|_F^2 +  \|R_w - Q^T B\|_F^2 \\
& +  \|XQ^T - YB^T\|_F^2
\end{aligned}
}
\end{equation}
\normalsize

This term can be regarded as an alignment constraint, aiming to ensure the consistency between the feature space and the label space in the low-dimensional representation. Specifically, \( X Q^T \) represents the low-dimensional mapping of features in the shared semantic space \( V \), while \( Y B^T \) represents the low-dimensional mapping of labels in the same space \( V \). Furthermore, to preserve the associations between features and labels, similar feature samples and label samples should be mapped to locations close to each other in the low-dimensional space. This ensures the learning of a common low-dimensional space that effectively captures the relationship between features and labels.

\begin{equation}
\small
{
\begin{aligned}
\min_{V, Q, B}\ &  \|X - VQ\|_F^2 +  \|Y - VB\|_F^2 +  \|R_w - Q^T B\|_F^2 \\
& +  \|XQ^T - YB^T\|_F^2 +  \|Q^T B\|_{2,1}  +  \|V\|_F^2
\end{aligned}
}
\end{equation}
\normalsize

The Frobenius norm of \( V \) imposes a sparsity constraint in the low-dimensional shared semantic space, reducing redundancy and unnecessary complexity, thereby enhancing the model's ability to capture important low-dimensional relationships between features and labels. Moreover, incorporating this term as a regularization on \( Q^T B \) effectively controls the model complexity and reduces the risk of overfitting. By smoothing the solution space, it further improves the generalization ability of the model. Additionally, computing the row sum of \( Q^T B \) provides a measure of feature importance.

Thus, our final objective function is formulated as follows:
\begin{equation}
\small
{
\begin{aligned}
\min_{V, Q, B} \ & \alpha \|X - VQ\|_F^2 + \beta \|Y - VB\|_F^2 + \gamma \|R_w - Q^T B\|_F^2 \\
& + \delta \|XQ^T - YB^T\|_F^2 + \epsilon \|Q^T B\|_{2,1}   +  \|V\|_F^2\\
\text{s.t.} &\ \{V, Q, B\} \geq 0 
\end{aligned}
}
\end{equation}

\section{Solution strategy}
This section presents the optimization method for solving the proposed objective function and provides a convergence analysis of the objective function under this method.
\subsection*{Optimization scheme}
Because the objective function includes an \( l_{2,1} \)-norm regularization term, which is non-smooth and cannot be directly solved, and is also non-convex with respect to the variables  \( V \), \( Q \), and \( B \), it presents a significant challenge \cite{boyd2004convex}. Theoretically, this is due to the fact that the Hessian matrix, composed of the second-order partial derivatives of the objective function, is not positive semi-definite. Therefore, to optimize the objective function, we have designed a relaxation update method based on the alternating-multiplier-based scheme \cite{nie2010efficient} to achieve the global optimum of the objective function.
\begin{equation}
\small
{
\begin{aligned}
\min_{V, Q, B} \ & \alpha \|X - VQ\|_F^2 + \beta \|Y - VB\|_F^2 + \gamma \|R_w - Q^T B\|_F^2 \\
& + \delta \|XQ^T - YB^T\|_F^2 +  \|V\|_F^2 + 2\epsilon  \operatorname{Tr}(W^T DW) 
\end{aligned}
}
\end{equation}

We use \( 2 \operatorname{Tr}(W^T D W) \) to approximate \( \|W\|_{2,1} \), where \( W = Q^T B \). Here, \( D \) is a diagonal matrix whose elements are iteratively computed during the method's execution. The parameter \( c \) is introduced to mitigate perturbations in non-differentiable problems. The elements of \( D \) are defined as:
\begin{equation}
\small
D_{ii} = \frac{1}{2\sqrt{W_i^T W_i + c }}, \quad (c \to 0)
\end{equation}

To integrate non-negative constraint conditions into the objective function, we introduce Lagrangian multipliers  \(\psi\), \(\varphi\), and \(\mu\) to constrain  \(V\), \(Q\), and \(B\) respectively. Specifically,  \(\psi \in \mathbb{R}_{+}^{n \times k}\), \(\varphi \in \mathbb{R}_{+}^{k \times d}\), and \(\mu \in \mathbb{R}_{+}^{k \times c}\). Consequently, the original function (14) is equivalent to the following function:
\begin{equation}
\small
{
\begin{aligned}
\min_{V, Q, B} \ & \alpha \|X - VQ\|_F^2 + \beta \|Y - VB\|_F^2 + \gamma \|R_w - Q^T B\|_F^2 \\
& + \delta \|XQ^T - YB^T\|_F^2 +  \|V\|_F^2 + 2\epsilon \operatorname{Tr}(W^T DW) \\
& - \operatorname{Tr}(\psi V^T) - \operatorname{Tr}(\varphi Q^T) - \operatorname{Tr}(\mu B^T)
\end{aligned}
}
\end{equation}

Transform the squared Frobenius norm into the form of a matrix trace:

\begin{equation}
\begin{array}{rl}
\Theta = {} 
& \alpha \operatorname{Tr}\big((X - VQ)^T (X - VQ)\big) \\
& + \beta \operatorname{Tr}\big((Y - VB)^T (Y - VB)\big) \\
& + \gamma \operatorname{Tr}\big((R_w - Q^T B)^T (R_w - Q^T B)\big) \\
& + \delta \operatorname{Tr}\big((XQ^T - YB^T)^T (XQ^T - YB^T)\big) \\
& + \operatorname{Tr}(V^T V) + 2\epsilon \operatorname{Tr}(W^T DW) \\
& - \operatorname{Tr}(\psi V^T) - \operatorname{Tr}(\varphi Q^T) - \operatorname{Tr}(\mu B^T).
\end{array}
\end{equation}

We can obtain the following expressions by differentiating with respect to \( V \), \( Q \), and \( B \):
\begin{equation}
\small{
\left\{
\begin{aligned}
\frac{\partial \Theta}{\partial V} &= 2 (\alpha VQQ^T + \beta VBB^T + V- \alpha XQ^T  - \beta YB^T) - \psi \\
\frac{\partial \Theta}{\partial Q} &= 2 (\alpha V^T VQ + \gamma B B^T Q + \epsilon B B^T Q D  + \delta Q X^T X \\
& \quad - \alpha V^T X - \gamma B R_w^T  - \delta B Y^T X) - \varphi \\
\frac{\partial \Theta}{\partial B} &= 2 (\beta V^T VB + \gamma Q Q^T B + \epsilon Q D Q^T B + \delta B Y^T Y \\
& \quad   - \beta V^T Y - \gamma Q R_w  - \delta Q X^T Y) - \mu
\end{aligned}
\right.
}
\end{equation}

By employing the Karush-Kuhn-Tucker (KKT) conditions for optimization, we derive the following set of equations:
\begin{equation}
\small{
\left\{
\begin{aligned}
&(\alpha VQQ^T + \beta VBB^T + V  - \alpha XQ^T - \beta YB^T)  \circ V = 0\\
&(\alpha V^T V Q + \gamma B B^T Q + \epsilon B B^T Q D  + \delta Q X^T X  \\
&\quad - \alpha V^T X - \gamma B R_w^T - \delta B Y^T X) \circ Q = 0 \\[1ex]
&(\beta V^T V B + \gamma Q Q^T B + \epsilon Q D Q^T B  + \delta B Y^T Y   \\
&\quad  - \beta V^T Y - \gamma Q R_w - \delta Q X^T Y) \circ B = 0
\end{aligned}
\right.
}
\end{equation}
\normalsize

In the case of multiple variables with constraints on each variable, we alternately update each variable to ensure that each optimization step adheres to the constraint conditions. We take the partial derivatives with respect to  \( V \), \( Q \), and \( B \) respectively.
\begin{equation}
\small
 V \leftarrow V \odot \frac{\alpha XQ^T + \beta YB^T}{\alpha VQQ^T + \beta VBB^T + V }
\end{equation}
\begin{equation}
\small
 Q \leftarrow Q \odot \frac{\alpha V^T X + \gamma B R_w^T + \delta B Y^T X}{\alpha V^T V Q + \gamma B B^T Q + \epsilon B B^T Q D + \delta Q X^T X}
\end{equation}
\begin{equation}
\small
 B \leftarrow B \odot \frac{\beta V^T Y + \gamma Q R_w + \delta Q X^T Y}{\beta V^T V B + \gamma Q Q^T B + \epsilon Q D Q^T B + \delta B Y^T Y }
\end{equation}
\normalsize % 恢复正常字体大小

After obtaining each parameter of the objective function, we can rank all features of \(X\) in descending order based on the values of \(\| (Q^\top B)_i \|_2\) (for \(i = 1, \ldots, d\)).

\begin{table}[t]
\centering
\scriptsize{
\setlength{\tabcolsep}{2.5pt} % 调整列间距，默认值为6pt，可以缩小到适应半栏
\renewcommand{\arraystretch}{1.0} % 调整行间距
\begin{tabular}{l:llllll} % 使用现有的 ":" 虚线格式
\toprule
Dataset    & \#Training set & \#Test set & \#Features & \#Labels & \#Distinct  & \#Domain   \\ 
\hdashline[0.5pt/1pt] % 调整虚线样式
Arts        & 2000 & 3000   & 462        & 26     & $321\pm139$ & Web text
 \\ 
Business    & 2000 & 3000   & 438        & 30     & $321\pm139$ & Web text
 \\ 
Education   & 2000 & 3000   & 550        & 33     & $321\pm139$ & Web text
 \\ 
Health      & 2000 & 3000   & 612        & 32     & $321\pm139$ & Web text
 \\ 
Yeast       & 1500 & 917   & 103        & 14     & 198 & Biology \\ 
Flags       & 129  & 65    & 19         & 7      & 54  & Image \\ 
Emotions    & 391  & 202   & 72         & 6      & 27  & Music  \\ 
\hdashline[0.5pt/1pt] % 添加底部虚线
\end{tabular}
}
\caption{Elaborated information regarding the experimental datasets.}
\end{table}

\section{Experiments}
\begin{table*}[!t]
    \centering
    \setlength{\tabcolsep}{1mm}
    \small
    { % 缩放表格到文本宽度
    \begin{tabular}{l*{9}{c}}
        \toprule
        Dataset & GRW-SCMF & PPT+MI & PPT+CHI & MIFS & LRFS & RALM-FS & SSFS & LRDG \\
        \toprule
        \multicolumn{9}{c}{Micro-$F_1$} \\ % 修改此行
        \midrule
        Flags & \textbf{0.7403$\pm$0.010}&0.6646$\pm$0.040 & 0.6632$\pm$0.038 & 0.7225$\pm$0.050 & 0.6659$\pm$0.041 & 0.6315$\pm$0.048 & 0.6520$\pm$0.034 & {0.7040$\pm$0.047} \\
        Emotions &  \textbf{0.5833$\pm$0.066} & 0.3442$\pm$0.175 & 0.2730$\pm$0.138 & 0.0684$\pm$0.059 & 0.2257$\pm$0.185 & 0.0082$\pm$0.030 & 0.4215$\pm$0.117 & {0.3223$\pm$0.147} \\
        Yeast & \textbf{0.5900$\pm$0.028} & 0.5568$\pm$0.028 & 0.5622$\pm$0.032 & 0.5659$\pm$0.028 & 0.5523$\pm$0.030 & 0.5590$\pm$0.029 & 0.5395$\pm$0.033 & {0.5677$\pm$0.038} \\
        Arts & \textbf{0.2823$\pm$0.061} &   0.0904$\pm$0.053 & 0.0981$\pm$0.055 & 0.1391$\pm$0.078 & 0.1041$\pm$0.045 & 0.1018$\pm$0.061 & 0.1878$\pm$0.092 & {0.0584$\pm$0.024} \\
        Business & \textbf{0.6986$\pm$0.011} & 0.6729$\pm$0.004 & 0.6725$\pm$0.004 & 0.6835$\pm$0.008 & 0.6796$\pm$0.005 & 0.6696$\pm$0.001 & 0.6817$\pm$0.010 & 0.6723$\pm$0.005 \\
        Education & \textbf{0.3135$\pm$0.060}   & 0.1237$\pm$0.083 & 0.1199$\pm$0.085 & 0.0733$\pm$0.059 & 0.1495$\pm$0.081 & 0.1934$\pm$0.056 & 0.2500$\pm$0.084 & {0.2283$\pm$0.086} \\
	   Health & \textbf{0.5549$\pm$0.033} & 0.4017$\pm$0.074 & 0.3993$\pm$0.075 & 0.4681$\pm$0.080 & 0.4754$\pm$0.037 & 0.5157$\pm$0.044 & 0.4999$\pm$0.102  & 0.4419$\pm$0.065 \\
        \toprule
        \multicolumn{9}{c}{Macro-$F_1$} \\ % 修改此行
        \midrule
        Flags & \textbf{0.5992$\pm$0.015} & 0.5135$\pm$0.037 & 0.5174$\pm$0.046 & 0.5697$\pm$0.100& 0.5106$\pm$0.044 & 0.4872$\pm$0.044 & 0.4987$\pm$0.048 & {0.5486$\pm$0.089} \\
        Emotions & \textbf{0.5213$\pm$0.068} & 0.2825$\pm$0.145 & 0.1925$\pm$0.118 & 0.0409$\pm$0.037 & 0.1660$\pm$0.155 & 0.0051$\pm$0.018 & 0.2521$\pm$0.070 & {0.2798$\pm$0.142} \\
        Yeast & \textbf{0.2830$\pm$0.038} & 0.2337$\pm$0.038 & 0.2435$\pm$0.046 & 0.2511$\pm$0.039 & 0.2280$\pm$0.045 & 0.2397$\pm$0.042 & 0.2084$\pm$0.044 & {0.2504$\pm$0.050} \\
        Arts & \textbf{0.1162$\pm$0.030} & 0.0373$\pm$0.022 & 0.0397$\pm$0.023 & 0.0550$\pm$0.034 & 0.0433$\pm$0.019 & 0.0385$\pm$0.026 & 0.0790$\pm$0.038 & 0.0222$\pm$0.009 \\
        Business & \textbf{0.1086$\pm$0.026} & 0.0385$\pm$0.008 & 0.0395$\pm$0.008 & 0.0564$\pm$0.011 & 0.0547$\pm$0.007 & 0.0320$\pm$0.001 & 0.0546$\pm$0.014 & 0.0424$\pm$0.010 \\
        Education & \textbf{0.0855$\pm$0.019}  & 0.0342$\pm$0.025 & 0.0345$\pm$0.026 & 0.0195$\pm$0.017 & 0.0485$\pm$0.028 & 0.0525$\pm$0.014 & 0.0647$\pm$0.020 & 0.0632$\pm$0.027 \\
        Health & \textbf{0.1670$\pm$0.042} & 0.1023$\pm$0.044 & 0.1039$\pm$0.043 & 0.1181$\pm$0.046 & 0.1454$\pm$0.038 & 0.1592$\pm$0.042 & 0.1658$\pm$0.063 & 
        0.1038$\pm$0.055 \\
        \bottomrule
    \end{tabular}
    } 
    \normalsize % 恢复正常字体大小
    \caption{The classification performance of all methodologies, evaluated using the SVM classifier, is provided in terms of Micro-$F_1$ and Macro-$F_1$ (mean$\pm$std).}
\end{table*}
\subsection*{Experimental Setup}
\noindent \textbf{Datasets.} To validate the effectiveness of the proposed method under complex relationships, our experimental evaluation employs multi-label datasets from the MULAN library \cite{tsoumakas2011mulan}, including the web text datasets Arts, Business, Education, and Health. Additionally, the datasets encompass the Emotions dataset for music genres, the Flags dataset for image data, and the Yeast dataset for biological data. The Arts, Business, Education, and Health datasets feature a large number of distinct label combinations, reflecting complex underlying relationships with the features. Meanwhile, the Emotions, Flags, and Yeast datasets highlight characteristics from different domains. Table 1 summarizes the characteristics of the datasets used in the experiments.

\bigskip

\noindent \textbf{Comparing Methods.} We compared a broad and representative set of multi-label feature selection methods: 
(1) \textbf{Statistical methods}: PPT+MI \cite{doquire2011feature} and PPT+CHI \cite{read2008pruned}, which perform feature selection based on mutual information and $\chi^2$ statistics, respectively; 
(2) \textbf{Latent representation learning}: LRDG \cite{zhang2024multi}, which incorporates dynamic graph constraints; 
(3) \textbf{Label relevance-based methods}: MIFS \cite{jian2016multi}, which selects features in a low-dimensional space, and LRFS \cite{zhang2019distinguishing}, which leverages conditional mutual information to capture label relationships; 
(4) \textbf{Sparse constraint-based methods}: RALM-FS \cite{cai2013exact}, which employs the $l_{2,0}$-norm for sparse solutions; 
(5) \textbf{Latent shared structure-based methods}: SSFS \cite{gao2021multilabel}, which models the shared latent structure between features and labels.
 
\bigskip

\noindent \textbf{Evaluating Methods.} Analogous to method \cite{hu2020multi}, we select the top 20\% of ranked features from each dataset, with a step size of 1\% (for the Flags dataset, all features were used as it only has 19 features). Three commonly used classifiers were employed: Linear SVM, k-Nearest Neighbors (k=3), and MLkNN (k=10). Specifically, both the Linear SVM and kNN classifiers provided Micro-$F_1$ and Macro-$F_1$ scores, while the MLkNN classifier yields Hamming Loss (HL) and Zero-One Loss (ZOL) \cite{kou2023novel,li2023multi}.
\bigskip

\noindent \textbf{Parameter Selection.} In this method, we employ Bayesian optimization to determine the optimal combination of parameters. The search range for the regularization parameters is set to \{0.01, 0.1, 0.3, 0.5, 0.7, 0.9, 1.0\}. For the random walk method, the range for $n_{walks}$ is set to \{100, 1000, 10000\}, the range for $walk\_length$ is set to \{10, 20, 30\}, and the ranges for $jump\_prob$ and $decay\_factor$ are both set to \{0.1, 0.2, 0.3, 0.4, 0.5, 0.6, 0.7, 0.8, 0.9\}.

\begin{table*}[!t]
    \centering
    \setlength{\tabcolsep}{1mm}
    \small{ % 缩放表格到文本宽度
     \begin{tabular}{l*{9}{c}} 
        \toprule
        Dataset & GRW-SCMF & PPT+MI & PPT+CHI & MIFS & LRFS & RALM-FS & SSFS & LRDG \\
        \toprule
        \multicolumn{9}{c}{Micro-$F_1$} \\ % 修改此行
        \midrule
        Flags & \textbf{0.6871$\pm$0.012} & 0.6150$\pm$0.011 & 0.6154$\pm$0.011 & 0.6615$\pm$0.030 & 0.6157$\pm$0.018 & 0.6057$\pm$0.012 & 0.6133$\pm$0.014 & 0.6632$\pm$0.037 \\
        Emotions & \textbf{0.5977$\pm$0.061} & 0.5324$\pm$0.038 & 0.5276$\pm$0.049 & 0.4877$\pm$0.074 & 0.4771$\pm$0.032 & 0.5492$\pm$0.075 & 0.3701$\pm$0.048 & 0.4452$\pm$0.069 \\
        Yeast & \textbf{0.5725$\pm$0.027} & 0.5494$\pm$0.025 & 0.5477$\pm$0.023 & 
        0.5547$\pm$0.021 & 0.5406$\pm$0.027 & 0.5477$\pm$0.024 & 0.3131$\pm$0.132 & 
        0.5553$\pm$0.032 \\
        Arts &  \textbf{0.2856$\pm$0.023} & 0.1812$\pm$0.037 & 0.1870$\pm$0.034 &
        0.2016$\pm$0.052 & 0.1963$\pm$0.027 & 0.1824$\pm$0.044 & 0.2300$\pm$0.049 & 0.1272$\pm$0.040 \\
        Business & \textbf{0.6745$\pm$0.011} & 0.6515$\pm$0.023 & 0.6556$\pm$0.023 & 0.6605$\pm$0.066 & 0.6700$\pm$0.009 & 0.6300$\pm$0.077 & 0.6626$\pm$0.020 & 
        0.6598$\pm$0.015 \\
        Education & \textbf{0.3122$\pm$0.027}  & 0.2301$\pm$0.043 & 0.2274$\pm$0.043 & 0.1827$\pm$0.055 & 0.2431$\pm$0.046 & 0.2605$\pm$0.051 & 0.2856$\pm$0.043 & 
        0.2769$\pm$0.054 \\
	   Health & \textbf{0.4878$\pm$0.012} & 0.3905$\pm$0.044 & 0.3870$\pm$0.045 & 
    0.4350$\pm$0.071 & 0.4365$\pm$0.020 & 0.4739$\pm$0.059 & 0.4537$\pm$0.052 & 
    0.4256$\pm$0.080 \\
        \toprule
        \multicolumn{9}{c}{Macro-$F_1$} \\ % 修改此行
        \midrule
        Flags & \textbf{0.5782$\pm$0.011} & 0.4480$\pm$0.011 & 0.4493$\pm$0.011 & 0.5265$\pm$0.078 & 0.4531$\pm$0.022 & 0.4386$\pm$0.011 & 0.4504$\pm$0.020 &
        0.5179$\pm$0.069 \\
        Emotions & \textbf{0.5824$\pm$0.064} & 0.5143$\pm$0.040 & 0.4997$\pm$0.051 & 
        0.4746$\pm$0.075 & 0.4551$\pm$0.030 & {0.5355$\pm$0.076} & 0.1751$\pm$0.028 & 
        0.4274$\pm$0.072 \\
        Yeast & \textbf{0.3718$\pm$0.030} & 0.3404$\pm$0.027 & 0.3376$\pm$0.025 & 
        0.3407$\pm$0.022 & 0.3287$\pm$0.028 & {0.3421$\pm$0.025} & 0.1380$\pm$0.049 & 
        0.3287$\pm$0.059 \\
        Arts & \textbf{0.1261$\pm$0.027}  & 0.0879$\pm$0.022 & 0.0912$\pm$0.022 & 
        0.0951$\pm$0.034 & 0.1006$\pm$0.025 & 0.0714$\pm$0.025 & 0.1186$\pm$0.035 &
        0.0534$\pm$0.016 \\
        Business & \textbf{0.1146$\pm$0.022} & 0.0810$\pm$0.019 & 0.0785$\pm$0.015 & 
        0.1075$\pm$0.023 & 0.0958$\pm$0.017 & 0.0654$\pm$0.018 & 0.0985$\pm$0.025 & 
        0.0931$\pm$0.026 \\
        Education & \textbf{0.1002$\pm$0.015}  & 0.0816$\pm$0.019 & 0.0829$\pm$0.022 & 0.0426$\pm$0.018 & 0.0870$\pm$0.020 & 0.0838$\pm$0.023 & 0.0959$\pm$0.020 & {0.0865$\pm$0.029} \\
	   Health & 0.1703$\pm$0.024 & 0.1329$\pm$0.029 & 0.1327$\pm$0.027 & 0.1570$\pm$0.041 & 0.1611$\pm$0.023 & \textbf{0.1859$\pm$0.032} & 0.1738$\pm$0.037  & {0.1302$\pm$0.053} \\
        \bottomrule
    \end{tabular}
    } 
    \normalsize % 恢复正常字体大小
    \caption{The classification performance of all methodologies, assessed using the 3NN classifier, is presented in terms of Micro-$F_1$ and Macro-$F_1$ (mean$\pm$std).}
\end{table*}

\begin{table}[h]
\centering
\begin{tabular}{c c c c c}
\toprule
{RW} & {FLA} & {Emotions} & {Yeast} & {Arts} \\
\midrule
 & $\checkmark$ & 0.5532 & 0.5808 & 0.2820 \\ 
$\checkmark$ &  & 0.5027 & 0.5846 & 0.2084 \\ 
$\checkmark$ & $\checkmark$ & 0.5833 & 0.5900 & 0.2823 \\
\bottomrule
\end{tabular}
\caption{Ablation experimental results.}
\end{table}

\begin{figure}[t!]
    \centering
    \begin{subfigure}[b]{0.38\linewidth}
        \centering
        \includegraphics[width=1\linewidth]{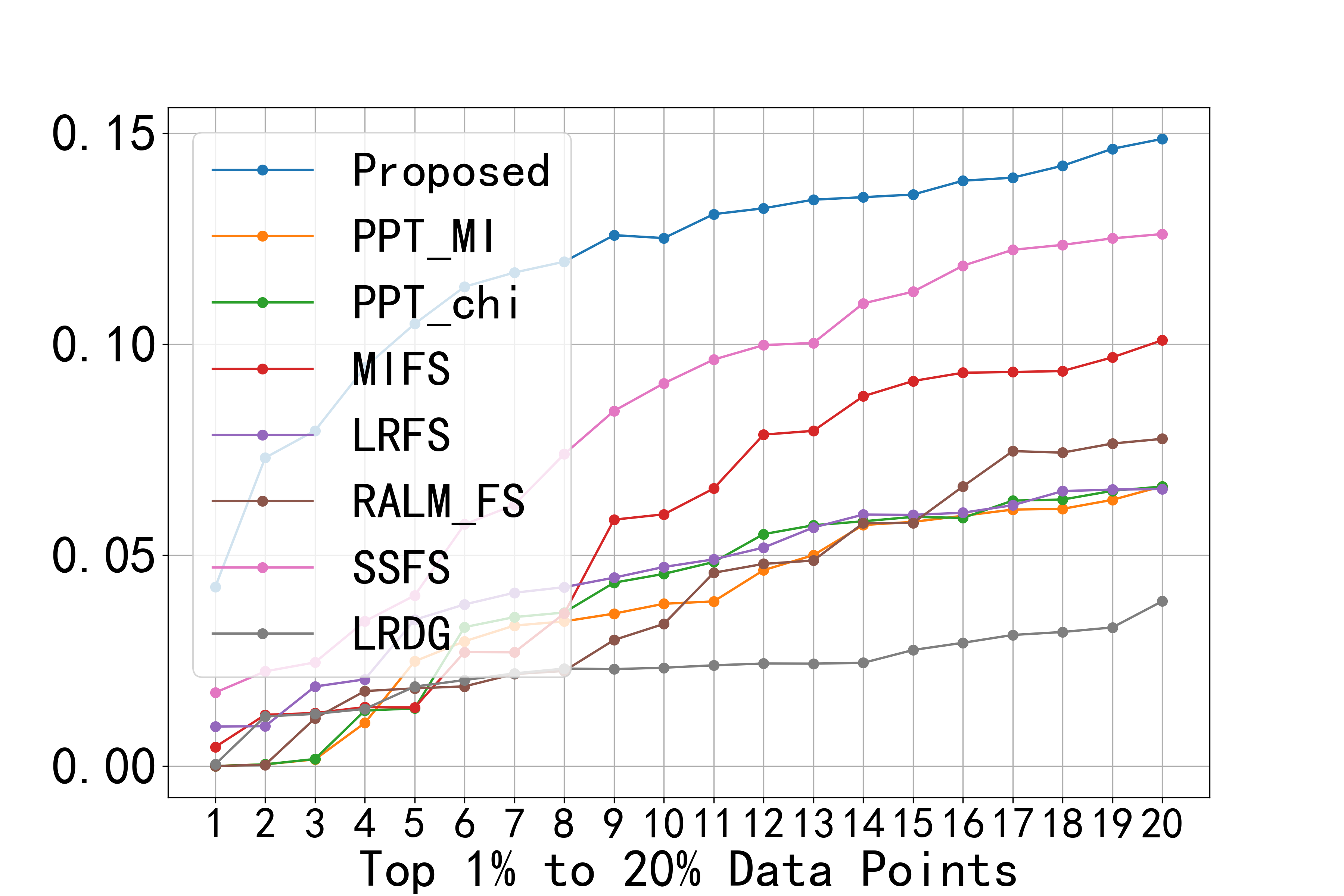}
        \caption{SVM in terms of Macro-$F_1$}
        \label{fig:f1_macro_scores_a_3d_bar}
    \end{subfigure}
    \hspace{0.05\linewidth}
    \begin{subfigure}[b]{0.38\linewidth}
        \centering
        \includegraphics[width=1\linewidth]{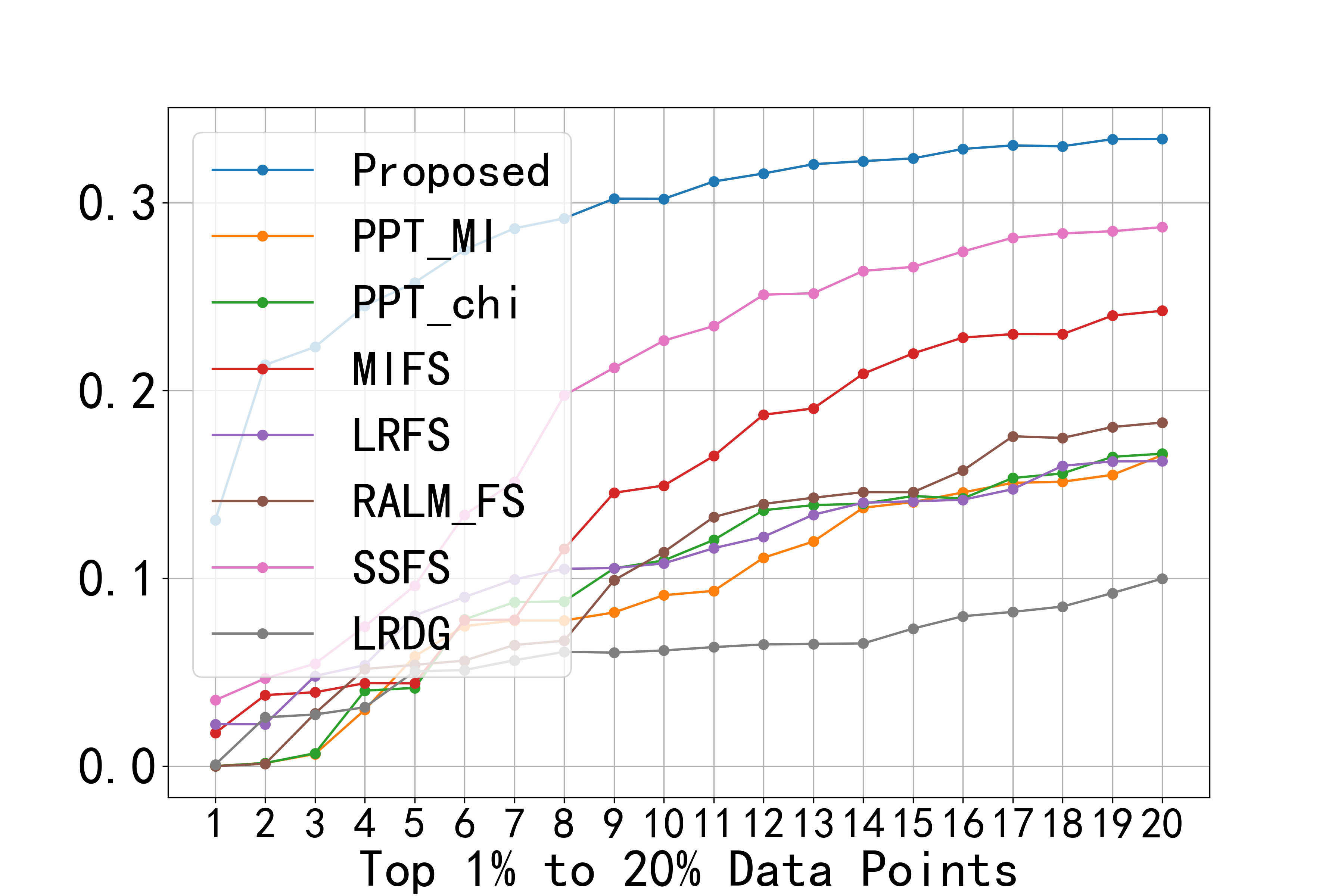}
        \caption{SVM in terms of Micro-$F_1$}
        \label{fig:f1_micro_scores_a_3d_bar}
    \end{subfigure}

    \begin{subfigure}[b]{0.38\linewidth}
        \centering
        \includegraphics[width=1\linewidth]{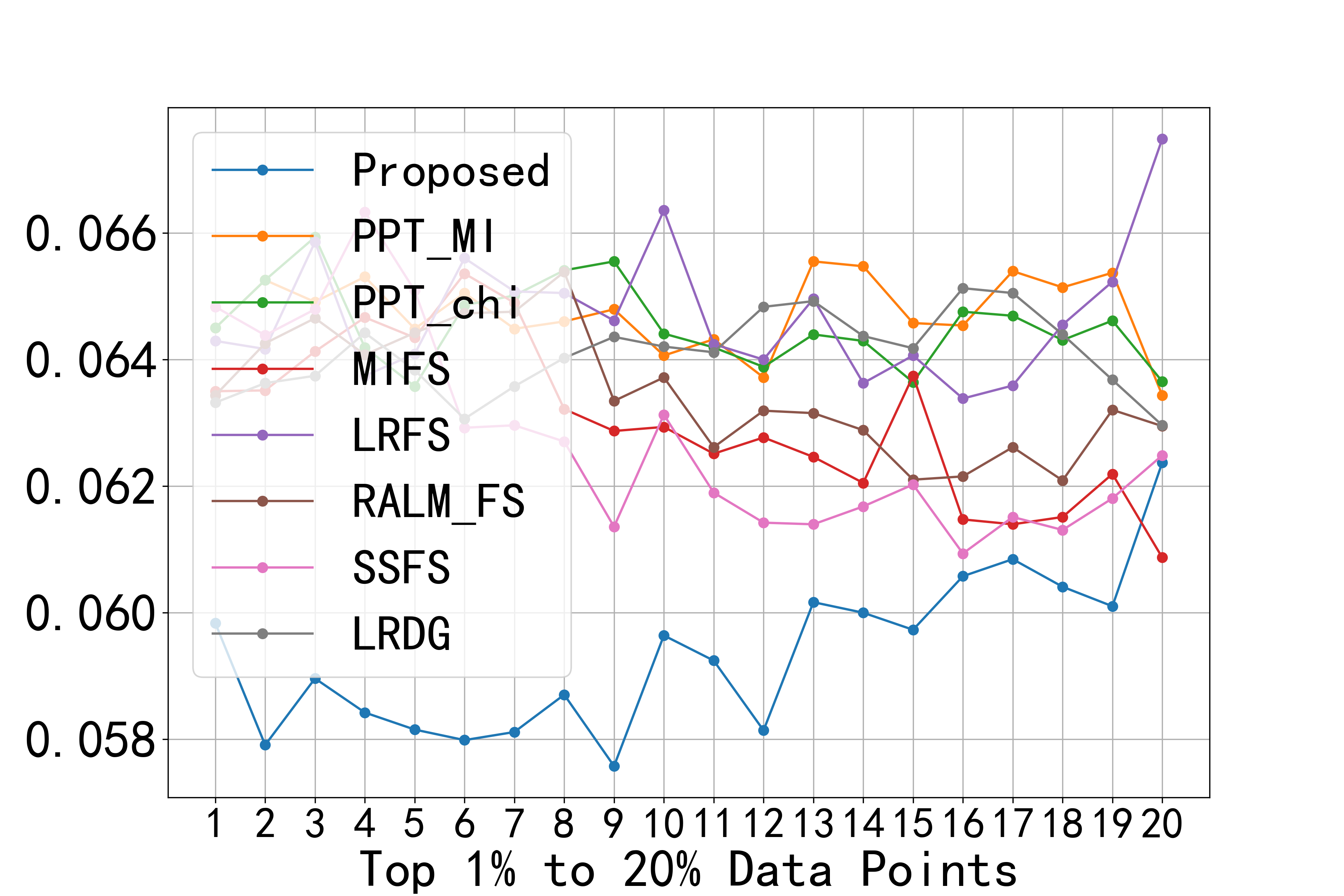}
        \caption{MLkNN in terms of HL}
        \label{fig:f1_macro_scores_b_3d_bar}
    \end{subfigure}
    \hspace{0.05\linewidth}
    \begin{subfigure}[b]{0.38\linewidth}
        \centering
        \includegraphics[width=1\linewidth]{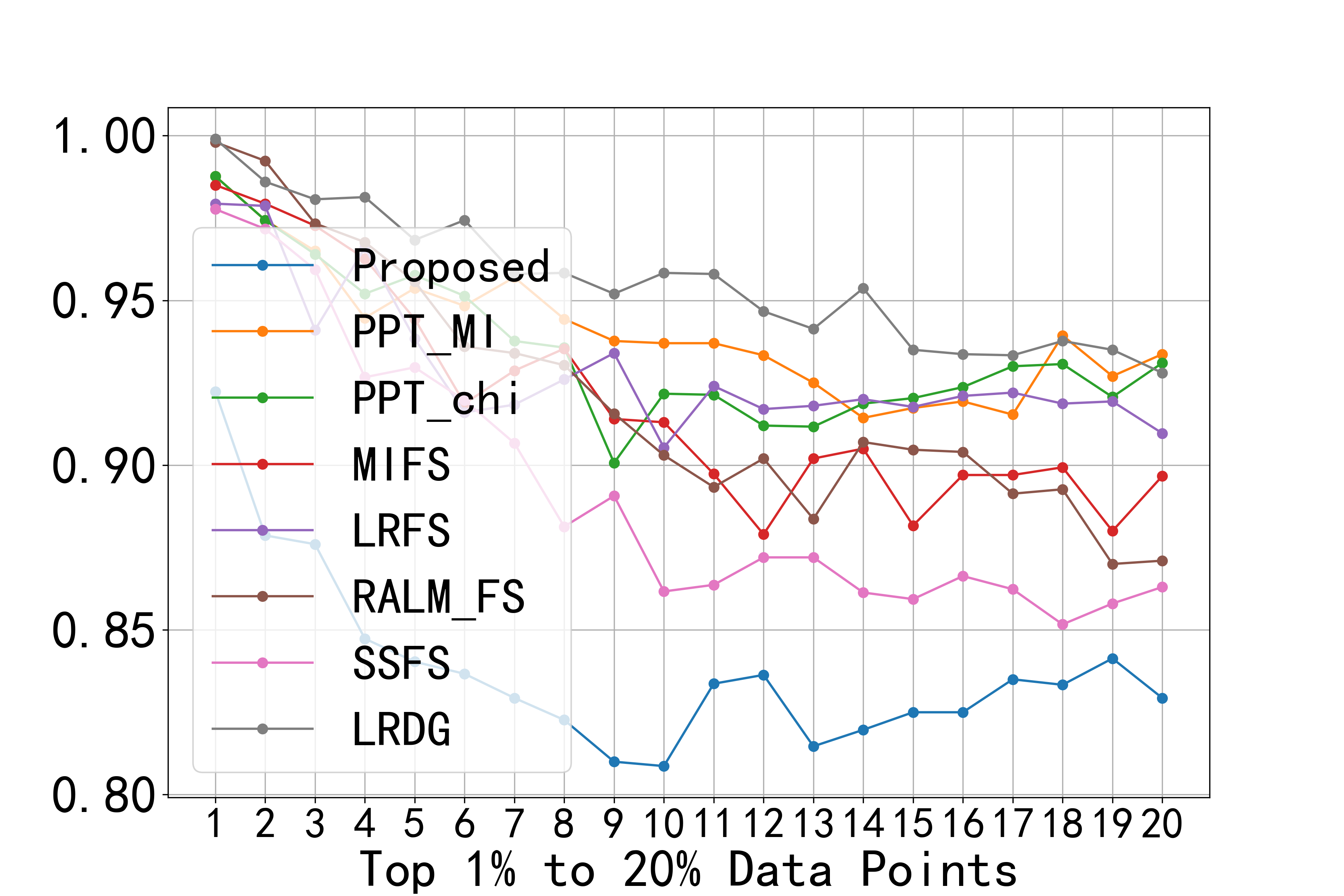}
        \caption{MLkNN in terms of ZOL}
        \label{fig:f1_micro_scores_b_3d_bar}
    \end{subfigure}
    \caption{Eight methods on Arts.}
    \label{fig:f1_scores_comparison_c}
\end{figure}

\begin{figure}[t!]
    \centering
    \begin{subfigure}[b]{0.32\linewidth}
        \centering
        \includegraphics[width=0.9\linewidth]{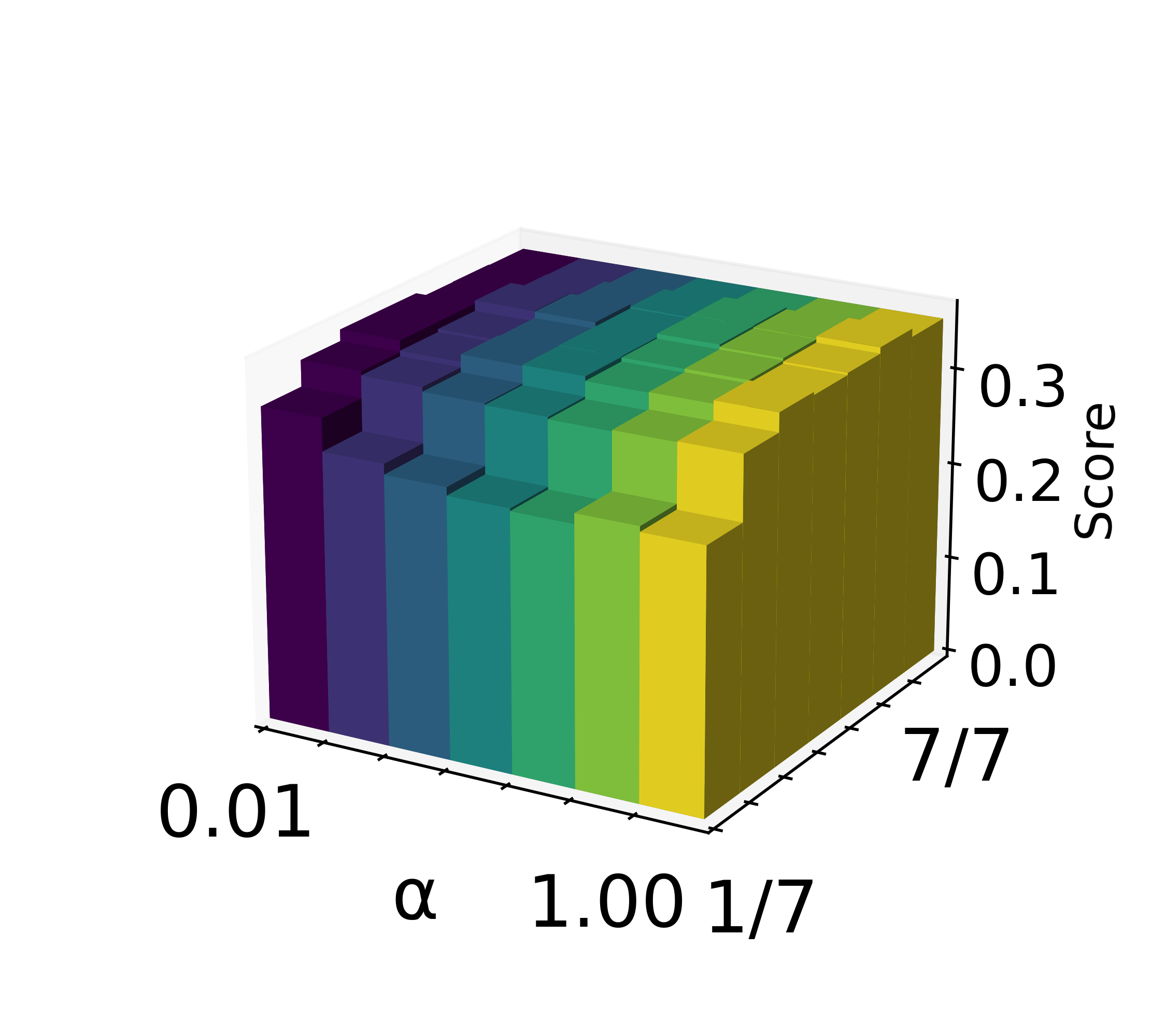}
        \caption{$\alpha$}
        \label{fig:f1_micro_scores_a_3d_bar}
    \end{subfigure}
    \hfill
    \begin{subfigure}[b]{0.32\linewidth}
        \centering
        \includegraphics[width=0.9\linewidth]{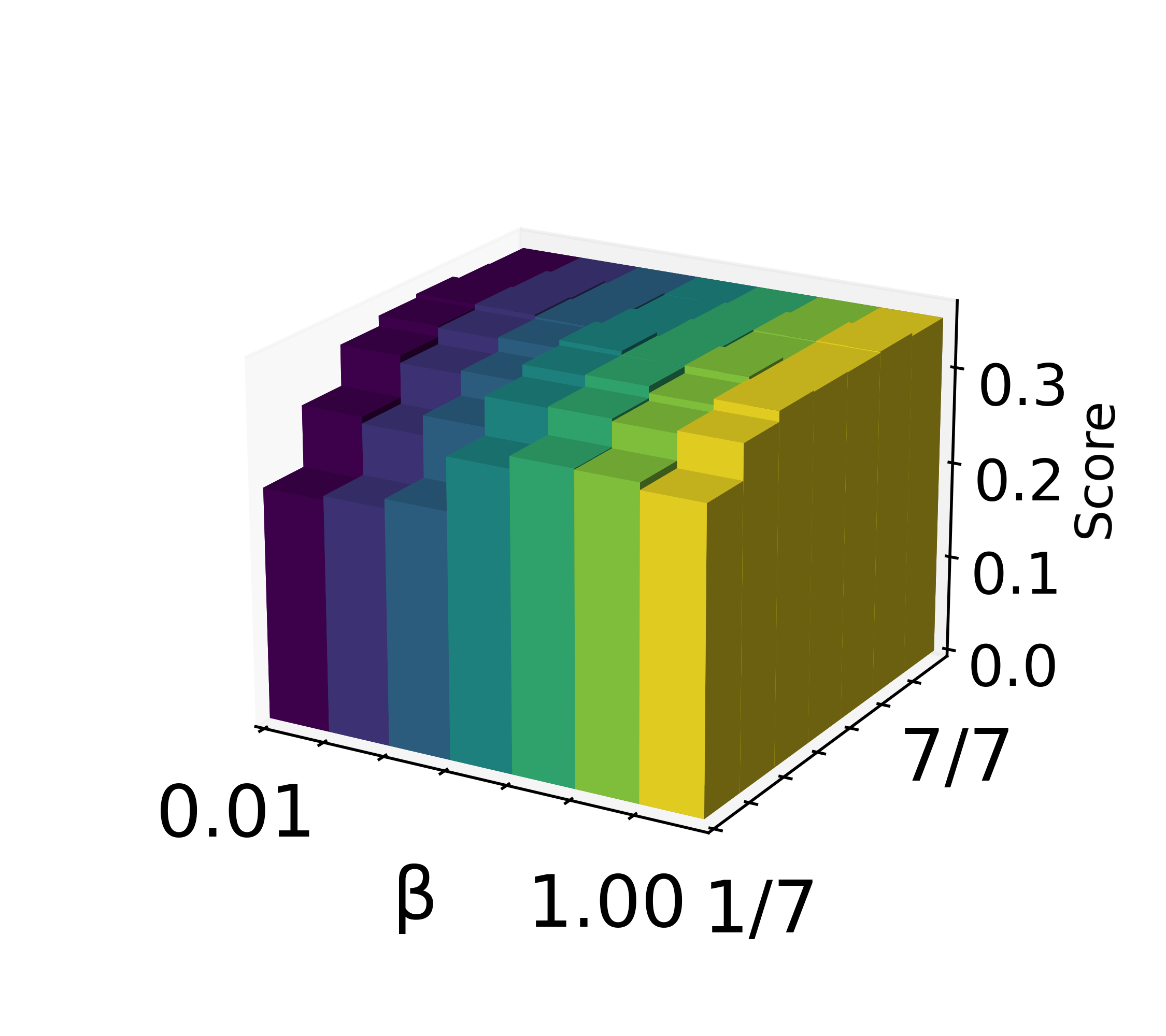}
        \caption{$\beta$}
        \label{fig:f1_micro_scores_b_3d_bar}
    \end{subfigure}
    \hfill
    \begin{subfigure}[b]{0.32\linewidth}
        \centering
        \includegraphics[width=0.9\linewidth]{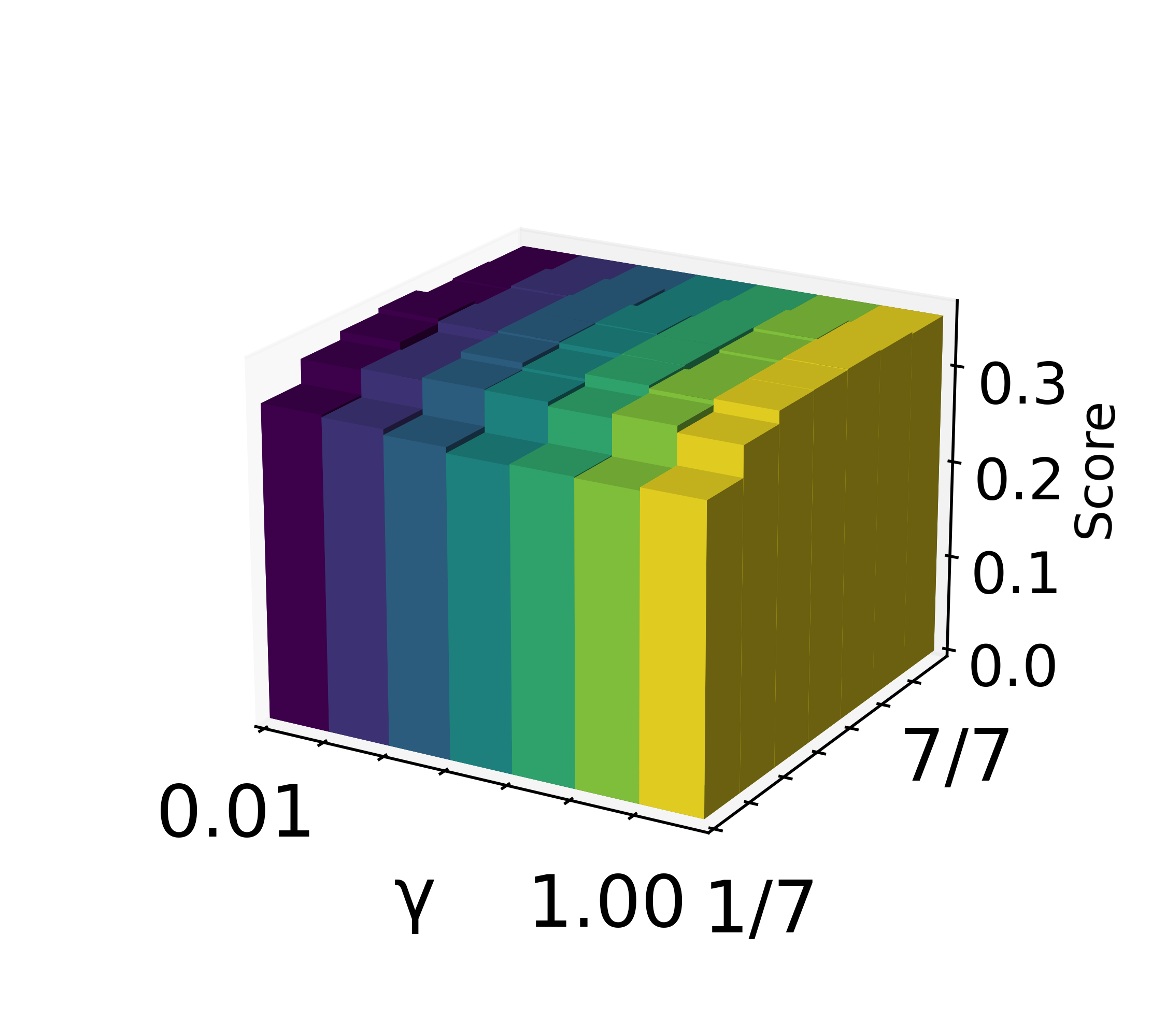}
        \caption{$\gamma$}
        \label{fig:f1_micro_scores_c_3d_bar}
    \end{subfigure}

    \begin{subfigure}[b]{0.32\linewidth}
        \centering
        \includegraphics[width=0.9\linewidth]{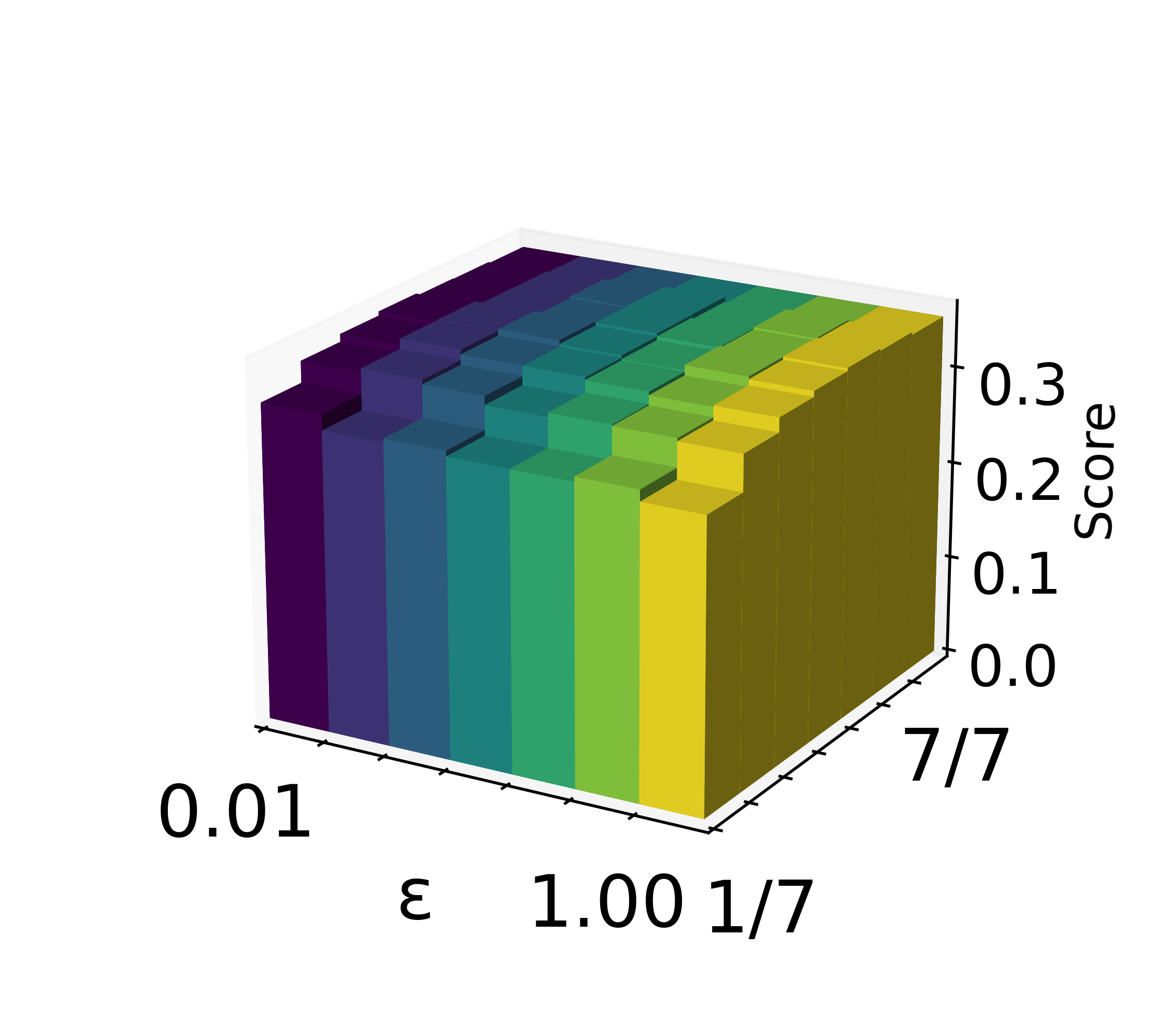}
        \caption{$\epsilon$}
        \label{fig:f1_micro_scores_e_3d_bar}
    \end{subfigure}
    \hfill
    \begin{subfigure}[b]{0.32\linewidth}
        \centering
        \includegraphics[width=0.9\linewidth]{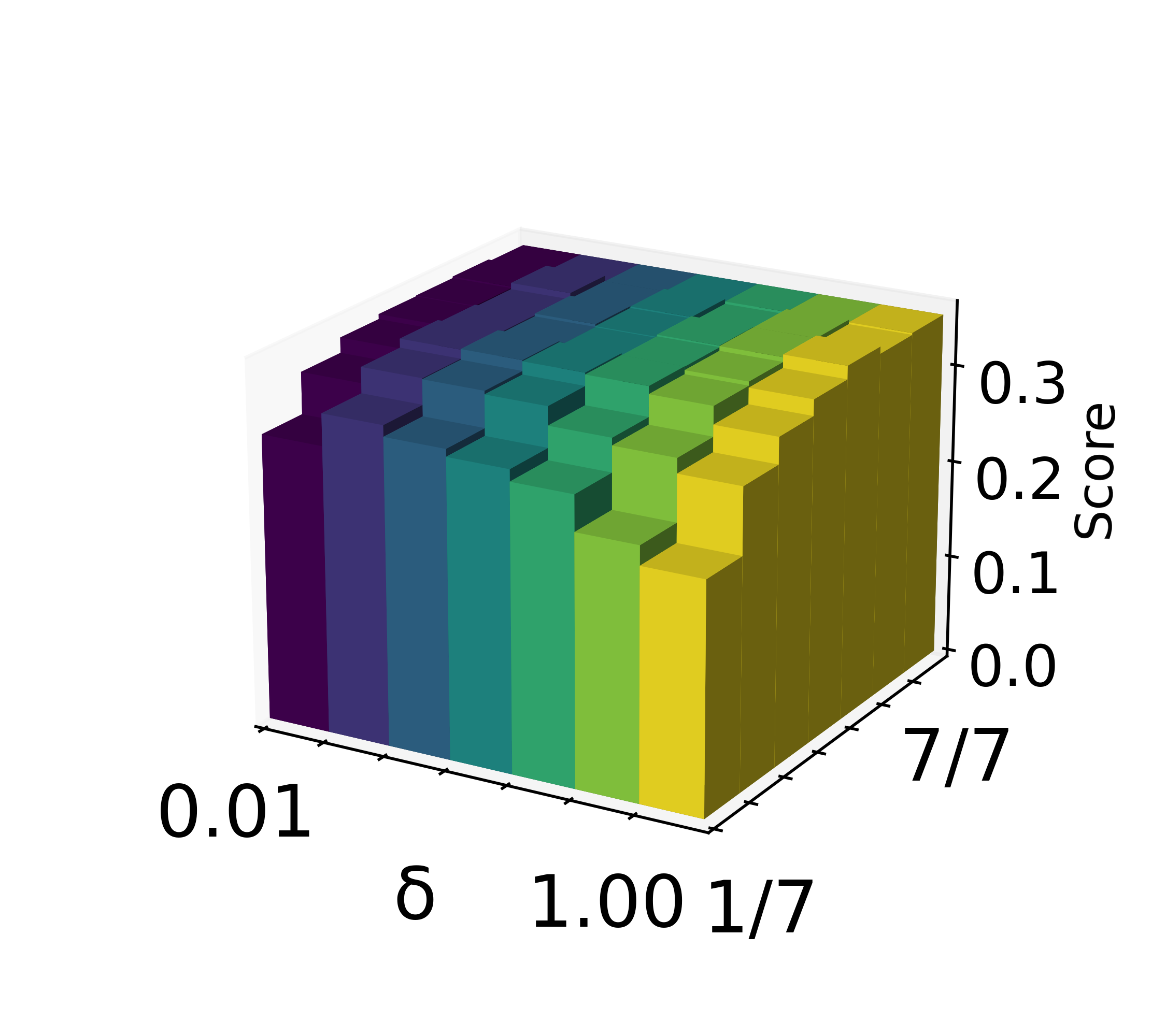}
        \caption{$\delta$}
        \label{fig:f1_micro_scores_d_3d_bar}
    \end{subfigure}
    
    \caption{Parameter sensitivity studies on the Arts dataset.}
    \label{fig:f1_scores_comparison}
\end{figure}
\subsection*{Experimental Results}
\noindent \textbf{Feature Selection Performance.} We selected seven benchmark multi-label datasets for our experiments and compared our method with seven state-of-the-art multi-label feature selection methods. The experimental results are presented in the tables, where our method outperforms the others.
In Tables 2 and 3, the best-performing method for each dataset is highlighted in bold. Clearly, our method is not limited to any specific type of dataset and demonstrates outstanding performance across image, music, and text datasets. Except for the Macro-$F_1$ score of the 3NN classifier on the Health dataset, where our method slightly underperforms compared to the RALM-FS method, our method outperforms the seven other methods in all other scenarios. In Figure 3, we present the performance of different methods on a specific dataset. From the early stages, the proposed method demonstrates its ability to select highly discriminative features. The outstanding results in HL and ZOL metrics validate its effectiveness in capturing both direct and indirect relationships between features and labels. Furthermore, the superior performance in Macro-$F_1$ and Micro-$F_1$ scores clearly highlights the efficacy of low-dimensional alignment.

\bigskip
\noindent \textbf{Parameter Analysis.} In our proposed method, there exist five weight parameters, $\alpha$ , 
 $\beta$ , $\gamma$ , $\delta$ and $\epsilon$, that influence performance outcomes. The figure~4 illustrates how these five parameters affect the Micro-$F_1$ scores computed using SVM on the Arts dataset. We initially divided the dataset into seven equal parts and then individually adjusted each parameter while keeping the other parameters fixed at 0.5 \cite{jian2018exploiting}, conducting a grid search over a predefined range. It is evident that the results exhibit an upward trend and become increasingly stable as the number of selected features increases. However, due to different parameter configurations, slight variations may occur under the same number of selected features, leading to minor discrepancies in the results. Therefore, we can infer that our method demonstrates stability, provided there is a sufficiently large number of instances.

\bigskip
\noindent \textbf{Ablation Study.} To evaluate the effectiveness of the newly introduced components, we conducted ablation experiments on the Emotions, Yeast, and Arts datasets, considering various combinations of these components, as shown in Table 4. In these experiments, the random walk component (RW) captures implicit indirect relationships through random walk, while the feature-label alignment component (FLA) aligns features and labels in a low-dimensional common space. The results demonstrate that removing either component adversely affects the SVM's Micro-$F_1$  performance, underscoring the importance of both introduced components. Therefore, it can be inferred that the contribution of each component is crucial for achieving optimal model performance.

\bigskip
\noindent \textbf{Convergence.} Figure~5 illustrates the convergence of our method on the Arts and Business datasets. To facilitate observation of oscillations, we set the initial point to approximately one. The chosen stopping criterion is: $\left| z_t - z_{t-1} \right| < c$ or $\left| \frac{z_t - z_{t-1}}{z_{t-1}} \right| < c$. The X-axis represents the number of iterations, and the Y-axis is similar to the stopping criterion but uses absolute values. It is evident from the figure that the optimization process converges quickly.

\begin{figure}[t!]
    \centering
    \begin{subfigure}[b]{0.48\linewidth}
        \centering
        \includegraphics[width=0.99\linewidth]{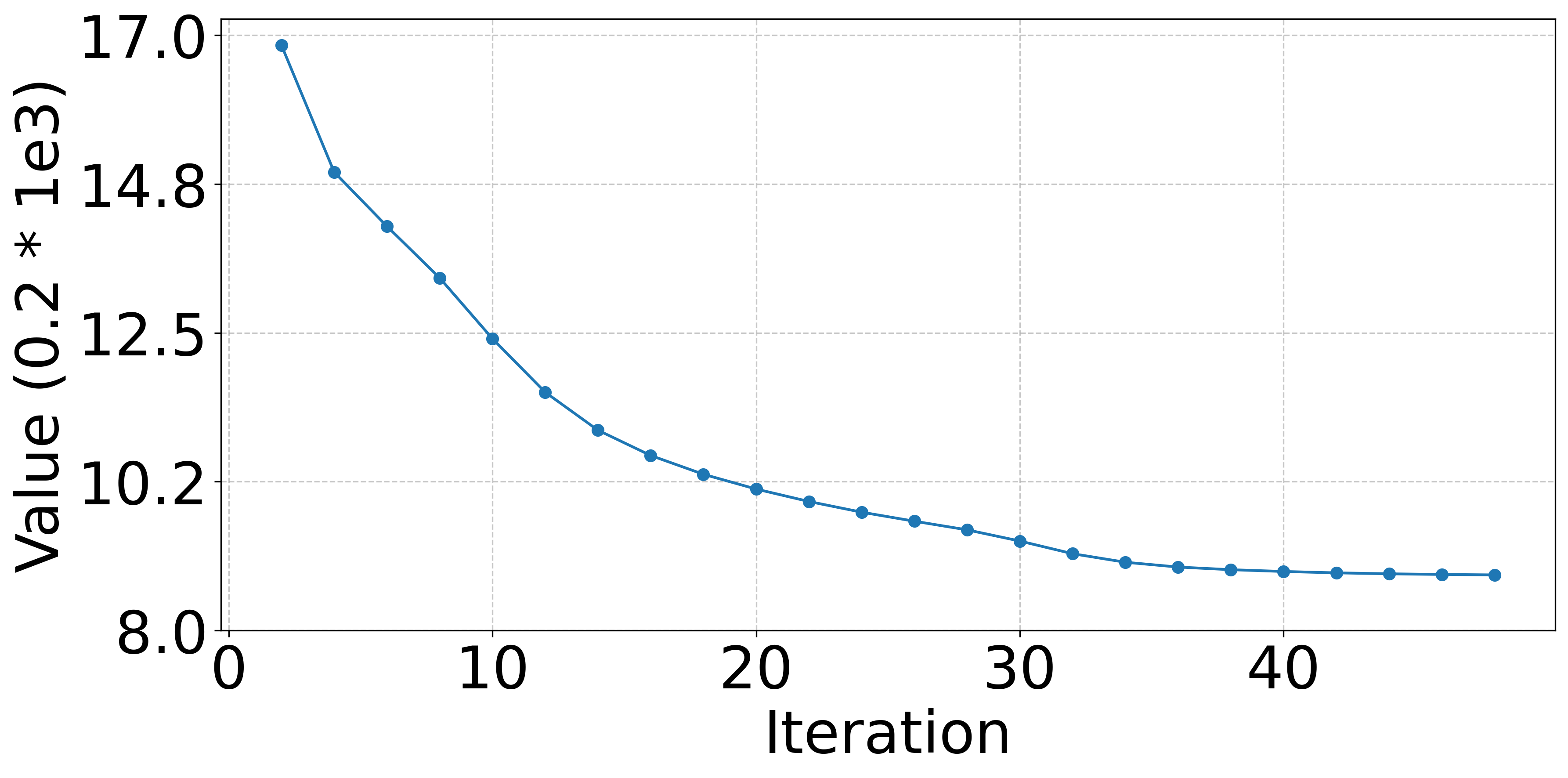}
        \caption{Arts}
        \label{fig:f1_macro_scores_b_3d_bar}
    \end{subfigure}
    \hspace{0.02\linewidth}
    \begin{subfigure}[b]{0.48\linewidth}
        \centering
        \includegraphics[width=0.99\linewidth]{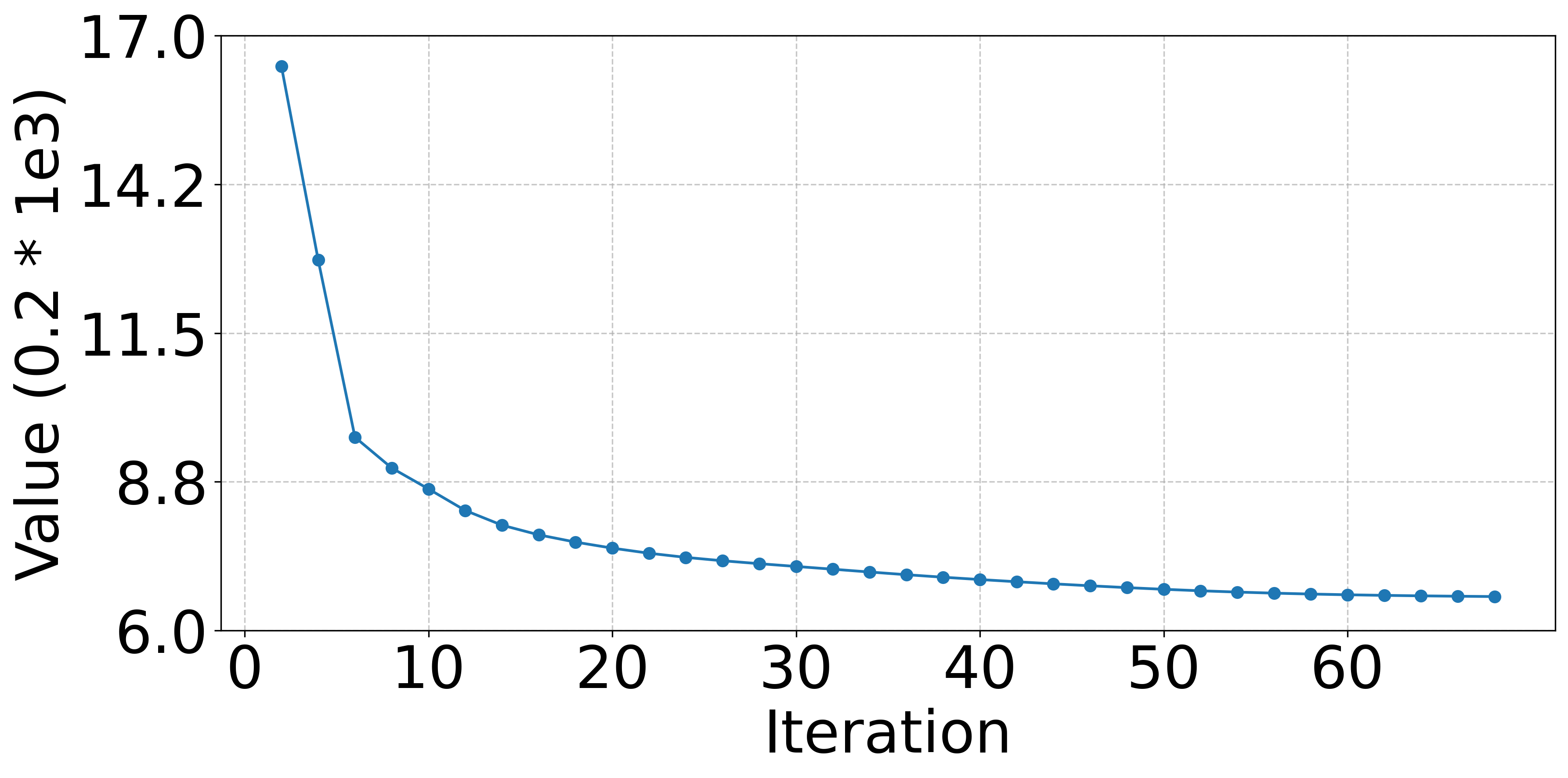}
        \caption{Business}
        \label{fig:f1_micro_scores_b_3d_bar}
    \end{subfigure}
    \caption{Convergence curves on Arts and Business datasets.}
    \label{fig:f1_scores_comparison_b}
\end{figure}

\section{Conclusion}

We propose a random walk method based on a feature-label composite graph and incorporate it into a multi-label feature selection method. Through random walks on the feature-label composite graph, we capture direct and indirect correlation between features and labels. Additionally, we leverage low-dimensional representation coefficients to align the low-dimensional variable space while preserving the manifold structure. Experimental results demonstrate the effectiveness and robustness of our method.

\clearpage
%% The file named.bst is a bibliography style file for BibTeX 0.99c
\section*{Acknowledgments}
This work was supported by the Science Foundation of Jilin Province of China under Grant YDZJ202501ZYTS286, and in part by Changchun Science and Technology Bureau Project under Grant 23YQ05.
\bibliographystyle{named}
\bibliography{ijcai25}

\end{document}